\documentclass[final]{cvpr}

\usepackage{times}
\usepackage{epsfig}
\usepackage{graphicx}
\usepackage{amsmath}
\usepackage{amssymb}

\usepackage{multirow}
\usepackage{makecell}
\usepackage{comment}
\usepackage[pagebackref=true,breaklinks=true,colorlinks,bookmarks=false]{hyperref}

\begin{document}

%%%%%%%%% TITLE
\title{Propagate Yourself: Exploring Pixel-Level Consistency for Unsupervised Visual Representation Learning}

\author{
Zhenda Xie\thanks{Equal Contribution. The work is done when Zhenda Xie and Yutong Lin are interns at Microsoft Research Asia.}\hspace{0.4mm} $^{13}$, Yutong Lin$^{*23}$, Zheng Zhang$^{3}$, Yue Cao$^{3}$, Stephen Lin$^{3}$, Han Hu$^{3}$ \\
  $^1$Tsinghua University \hspace{2mm} $^2$Xi'an Jiaotong University\\
  $^3$Microsoft Research Asia\\
  {\tt\small xzd18@mails.tsinghua.edu.cn \hspace{2mm} yutonglin@stu.xjtu.edu.cn} \\
  {\tt\small \{zhez,yuecao,stevelin,hanhu\}@microsoft.com} \\
}
\maketitle

%%%%%%%%% ABSTRACT
\begin{abstract}
Contrastive learning methods for unsupervised visual representation learning have reached remarkable levels of transfer performance. We argue that the power of contrastive learning has yet to be fully unleashed, as current methods are trained only on instance-level pretext tasks, leading to representations that may be sub-optimal for downstream tasks requiring dense pixel predictions. In this paper, we introduce pixel-level pretext tasks for learning dense feature representations. The first task directly applies contrastive learning at the pixel level. We additionally propose a pixel-to-propagation consistency task that produces better results, even surpassing the state-of-the-art approaches by a large margin. Specifically, it achieves \textbf{60.2} AP, \textbf{41.4 / 40.5} mAP and \textbf{77.2} mIoU when transferred to Pascal VOC object detection (C4), COCO object detection (FPN / C4) and Cityscapes semantic segmentation using a ResNet-50 backbone network, which are \textbf{2.6} AP, \textbf{0.8 / 1.0} mAP and \textbf{1.0} mIoU better than the previous best methods built on instance-level contrastive learning. Moreover, the pixel-level pretext tasks are found to be effective for pre-training not only regular backbone networks but also head networks used for dense downstream tasks, and are complementary to instance-level contrastive methods. These results demonstrate the strong potential of defining pretext tasks at the pixel level, and suggest a new path forward in unsupervised visual representation learning. Code is available at \url{https://github.com/zdaxie/PixPro}.

\end{abstract}

%%%%%%%%% BODY TEXT
\section{Introduction}

According to Yann LeCun, ``if intelligence is a cake, the bulk of the cake is unsupervised learning''. This quote reflects his belief that human understanding of the world is predominantly learned from the tremendous amount of unlabeled information within it. Research in machine intelligence has increasingly moved in this direction, with substantial progress in unsupervised and self-supervised learning\cite{wu2018memorybank,he2019moco,misra2019PIRL,chen2020simclr,tian2020infomin}. In computer vision, recent advances can largely be ascribed to the use of a pretext task called \emph{instance discrimination}, which treats each image in a training set as a single class and aims to learn a feature representation that discriminates among all the classes.

\begin{figure}
    \centering
    \includegraphics[width=0.9\linewidth]{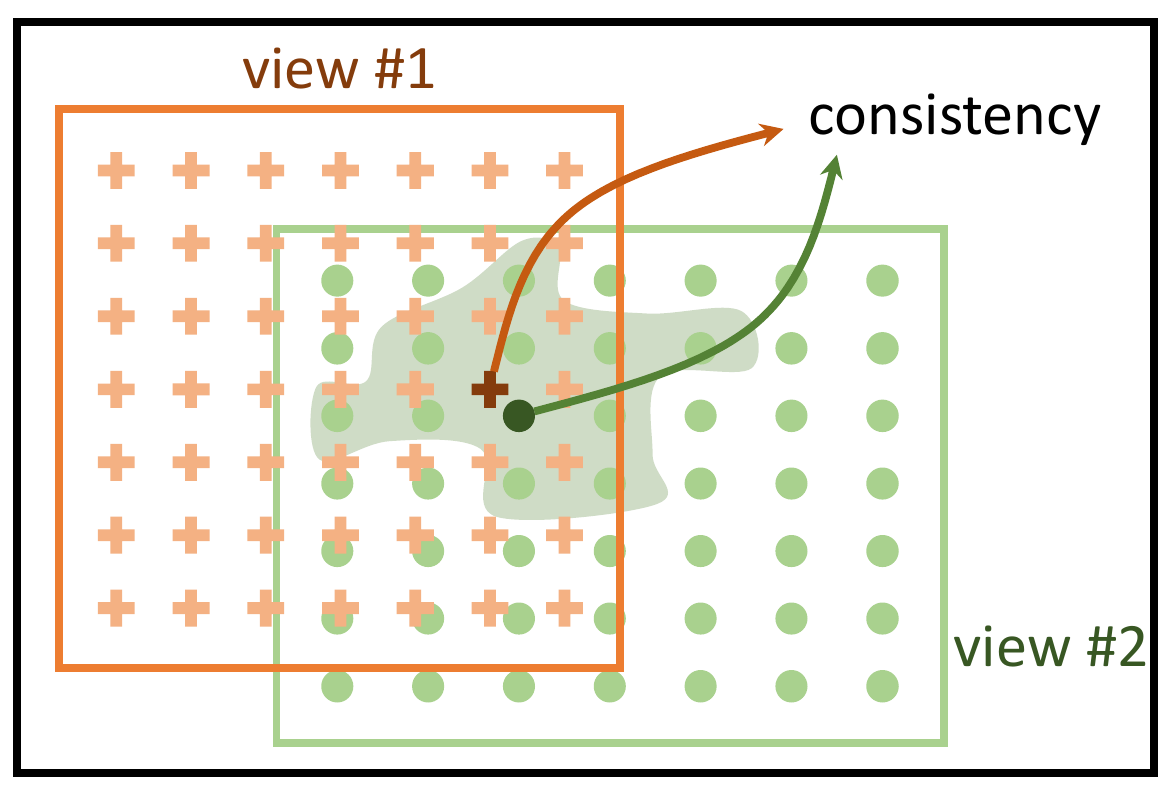}
    \caption{An illustration of the proposed \emph{PixPro} method, which is based on a \emph{pixel-to-propagation consistency} pretext task for pixel-level visual representation learning. In this method, two views are randomly cropped from an image (outlined in black), and the features from the corresponding pixels of the two views are encouraged to be consistent. For one of them, the feature comes from a regular pixel representation (illustrated as orange crosses). The other feature comes from a smoothed pixel representation (shown as green dots) built by propagating the features of similar pixels (illustrated as the light green region). Note that this hard selection of similar pixels is for illustration only. In implementation, all pixels on the same view will contribute to propagation, with the propagation weight of each pixel determined by its feature similarity to the center pixel. }
    \label{fig:teaser}
\end{figure}

Although self-supervised learning has proven to be remarkably successful, we argue that there remains significant untapped potential. The self-supervision that guides representation learning in current methods is based on image-level comparisons. As a result, the pre-trained representation may be well-suited for image-level inference, such as image classification, but may lack the spatial sensitivity needed for downstream tasks that require pixel-level predictions, e.g., object detection and semantic segmentation. How to perform self-supervised representation learning at the pixel level is a problem that until now has been relatively unexplored.

In this paper, we tackle this problem by introducing
pixel-level pretext tasks for self-supervised visual representation learning. Inspired by recent instance discrimination methods, our first attempt is to construct a pixel-level contrastive learning task, where each pixel in an image is treated as a single class and the goal is to distinguish each pixel from others within the image. Features from the same pixel are extracted via two random image crops containing the pixel, and these features are used to form positive training pairs. On the other hand, features obtained from different pixels are treated as negative pairs. With training data collected in this self-supervised manner, a contrastive loss is applied to learn the representation. We refer to this approach as \emph{PixContrast}.

In addition to this contrastive approach, we present a method based on \emph{pixel-to-propagation consistency}, where positive pairs are obtained by extracting features from the same pixel through two asymmetric pipelines instead. The first pipeline is a standard backbone network with a projection head. The other has a similar form but ends with a proposed pixel propagation module, which filters the pixel's features by propagating the features of similar pixels to it. This filtering introduces a certain smoothing effect, while the standard feature maintains spatial sensitivity. A difference of this method from the contrastive approach of \emph{PixContrast} is that it encourages consistency between positive pairs without consideration of negative pairs. While the performance of contrastive learning is known to be influenced heavily by how negative pairs are handled~\cite{he2019moco,chen2020simclr}, this issue is avoided in this consistency-based pretext task. Empirically, we find that this pixel-to-propagation consistency method, which we call \emph{PixPro}, significantly outperforms the \emph{PixContrast} approach over various downstream tasks.

Besides learning good pixel-level representations, the proposed pixel-level pretext tasks are found to be effective for pre-training on not only backbone networks but also head networks used for dense downstream tasks, contrary to instance-level discrimination where only backbone networks are pre-trained and transferred. This is especially beneficial for downstream tasks with limited annotated data, as all layers can be well-initialized. Moreover, the proposed pixel-level approach is complementary to existing instance-level methods, where the former is good at learning a spatially sensitive representation and the latter provides better categorization ability. A combination of the two methods capitalizes on both of their strengths, while also remaining computationally efficient in pre-training as they both can share a data loader and backbone encoders.

The proposed \emph{PixPro} achieves state-of-the-art transfer performance on common downstream benchmarks requiring dense prediction. Specifically, with a ResNet-50 backbone, it obtains 60.2 AP on Pascal VOC object detection using a Faster R-CNN detector (C4 version), 41.4 / 40.5 mAP on COCO object detection using a Mask R-CNN detector (both the FPN / C4 versions, $1\times$ settings), and 77.2 mIoU Cityscapes semantic segmentation using an FCN method, which are 2.6 AP, 0.8 / 1.0 mAP, and 1.0 mIoU better than the leading unsupervised/supervised methods. Though past evaluations of unsupervised representation learning have mostly been biased towards linear classification on ImageNet, we advocate a shift in attention to performance on downstream tasks, which is the main purpose of unsupervised representation learning and a promising setting for pixel-level approaches.

\section{Related Works}

\paragraph{Instance discrimination}  Unsupervised visual representation learning is currently dominated by the pretext task of instance discrimination, which treats each image as a single class and learns representations by distinguishing each image from all the others. This line of investigation can be traced back to~\cite{dosovitskiy2014exemplarcnn}, and after years of progress~\cite{wu2018memorybank,tian2019contrastive,henaff2019CPC,zhuang2019local,bachman2019learning,ye2019unsupervised}, transfer performance superior to supervised methods was achieved by MoCo~\cite{he2019moco} on a broad range of downstream tasks. After this milestone, considerable attention has been focused in this direction~\cite{chen2020simclr,tian2020infomin,cao2020pic,grill2020byol,caron2020swav}. While follow-up works have quickly improved the linear evaluation accuracy (top-1) on ImageNet-1K from about 60\%~\cite{he2019moco} to higher than 75\%~\cite{caron2020swav} using a ResNet-50 backbone, the improvements on downstream tasks such as object detection on Pascal VOC and COCO have been marginal.

Instead of using instance-level pretext tasks, our work explores pretext tasks at the pixel level for unsupervised feature learning. We focus on transfer performance to downstream tasks such as object detection and semantic segmentation, which have received limited consideration in prior research. We show that pixel-level representation learning can surpass the existing instance-level methods by a significant margin, demonstrating the potential of this direction.

\paragraph{Other pretext tasks using a single image} Aside from instance discrimination, there exist numerous other pretext tasks including context prediction~\cite{doersch2015context}, grayscale image colorization~\cite{zhang2016colorization}, jigsaw puzzle solving~\cite{noroozi2016jigsaw}, split-brain auto-encoding~\cite{zhang2017splitbrain}, rotation prediction~\cite{gidaris2018rotation}, learning to cluster~ \cite{caron2018deepcluster}, and missing part prediction~\cite{henaff2019CPC,trinh2019selfie,chen2020imagegpt}. Interest in these tasks for unsupervised feature learning has fallen off considerably due to their inferior performance and greater complexity in architectures or training strategies. Among these methods, the approach most related to ours is missing parts prediction~\cite{henaff2019CPC,trinh2019selfie,chen2020imagegpt}, which was inspired by successful pretext tasks in natural language processing~\cite{devlin2018bert,brown2020language}. 
Like our pixel-propagation consistency technique, such methods also operate locally. However, they either partition images into patches~\cite{trinh2019selfie,henaff2019CPC} or require special architectures/training strategies to perform well~\cite{henaff2019CPC,chen2020imagegpt}, while our approach directly operates on pixels and has no special requirements on the encoding networks. Training with our method is also simple, with few bells and whistles. More importantly, our approach achieves state-of-the-art transfer performance on the important dense prediction tasks of object detection and semantic segmentation.

\paragraph{Pixel-level self-supervised learning in videos or multi-images} Videos or multi-images naturally provide repetitive pixels on multiple views for correspondence learning~\cite{wang2019learning,li2019jointtask, jabri2020spacetime, kang2020pixellevel}. Since the ground-truth pixel correspondences on different images are unknown, these works usually form their pixel-level pretext task by a weakly cycle-consistency check between forward and backward association. Contrary to these works, we directly build the pixel-level correspondence pretext task by different views of a single image, where the ground-truth correspondence can be exactly computated. The utilizing of single images also enables us to leverage the large-scale image dataset for training (e.g. ImageNet-1K).

\paragraph{Concurrent/follow-on works of pixel-level learning using a single image} Concurrent to our work, there are some papers also exploring pixel-level pretext tasks for self-supervised representation learning~\cite{pinheiro2020unsupervised,chaitanya2020contrastive,wang2020DenseCL,xie2021detco}. Most of them are based on contrastive learning, where the negative pairs need to be carefully tuned. In our approach, while we set the contrastive learning as a direct extension of the instance discrimination methods, we additionally advocate a consistency pretext task for pixel-level representation learning. We also study the complementarity of the pixel-level learning and instance-level learning, the benefit by pre-training head networks, and the application to semi-supervised object detection. Our approach also achieves significantly better accuracy on benchmarks, particularly, on the Pascal VOC object detection benchmark.

\section{Method}

\subsection{Pixel-level Contrastive Learning}

\label{sec.baseline}

The state-of-the-art unsupervised representation learning methods are all built on the pretext task of instance discrimination. In this section, we show that the idea of instance discrimination can be also applied at the pixel level for learning visual representations that generalize well to downstream tasks. We adopt the prevalent contrastive loss to instantiate the pixel-level discrimination task, and call this method \emph{PixContrast}.

\begin{figure*}
    \centering
    \includegraphics[width=1.0\linewidth]{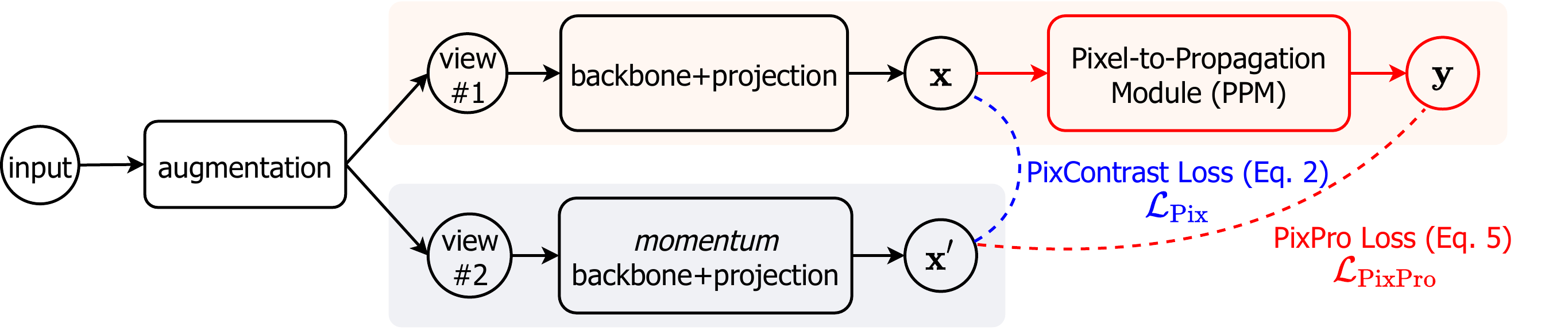}
    \caption{Architecture of the \emph{PixContrast} and \emph{PixPro} methods.}
    \label{fig:pix_architecture}
\vspace{-0.5em}
\end{figure*}

As done in most instance-level contrastive learning methods, \emph{PixContrast} starts by sampling two augmentation views from the same image. The two views are both resized to a fixed resolution (e.g., $224\times 224$) and are passed through a regular encoder network and a momentum encoder network~\cite{he2019moco,chen2020mocov2,grill2020byol} to compute image features. The encoder networks are composed of a backbone network and a projection head network, where the former could be any image neural network (we adopt ResNet by default), and the latter consists of two successive $1\times 1$ convolution layers (of 2048 and 256 output channels, respectively) with a batch normalization layer and a ReLU layer in-between to produce image feature maps of a certain spatial resolution, e.g., $7\times 7$. While previous methods compute a single image feature vector for each augmentation view, \emph{PixContrast} computes a feature map upon which pixel-level pretext tasks can be applied. The learnt backbone representations are then used for feature transfer. An illustration of the architecture is shown in Figure~\ref{fig:pix_architecture}.

\paragraph{Pixel Contrast} With the two feature maps computed from two views, we can construct pixel contrast pretext tasks for representation learning. Each pixel in a feature map is first warped to the original image space, and the distances between all pairs of pixels from the two feature maps are computed. The distances are normalized to the diagonal length of a feature map bin to account for differences in scale between the augmentation views, and the normalized distances are used to generate positive and negative pairs, based on a threshold $\mathcal{T}$:
\begin{equation}
\label{eq.assign_rule}
    A(i,j)=\left\{
    \begin{aligned}
    1, & \text{~~~~if~} \text{dist}(i,j) \leq \mathcal{T},\\0, & \text{~~~~if~} \text{dist}(i,j) > \mathcal{T},
    \end{aligned}
    \right.
\end{equation}
where $i$ and $j$ are pixels from each of the two views; $\text{dist}(i,j)$ denotes the normalized distance between pixel $i$ and $j$ in the original image space; and the threshold is set to $\mathcal{T} = 0.7$ by default.

Similar to instance-level contrastive learning methods, we adopt a contrastive loss for representation learning:
\begin{equation}
\label{eqn:cosine-softmax}
\small
    \mathcal{L}_\text{Pix}(i) =  - {\log \frac{\sum \limits_{j\in \Omega_p^i} e^ { { \cos \left( {\mathbf{x}_{i}},{\mathbf{x}'_{j}} \right)} /{\tau} }}{\sum \limits_{j\in \Omega_p^i} e^{ { \cos \left( {\mathbf{x}_{i}},{\mathbf{x}'_{j}} \right)}/ {\tau}}  + \sum\limits_{k\in \Omega_n^i}e^{ { \cos \left( {\mathbf{x}_{i}},{\mathbf{x}'_{k}} \right)}/ {\tau}}}},
\end{equation}
where $i$ is a pixel in the first view that is also located in the second view; $\Omega_p^i$ and $\Omega_n^i$ are sets of pixels in the second view assigned as positive and negative, respectively, with respect to pixel $i$; $\mathbf{x}_i$ and $\mathbf{x}_j'$ are the pixel feature vectors in two views; and $\tau$ is a scalar temperature hyper-parameter, set by default to $0.3$. The loss is averaged over all pixels on the first view that lie in the intersection of the two views. Similarly, the contrastive loss for a pixel $j$ on the second view is also computed and averaged. The final loss is the average over all image pairs in a mini-batch.

As later shown in the experiments, this direct extension of instance-level contrastive learning to the pixel level performs well in representation learning.

\subsection{Pixel-to-Propagation Consistency}
\label{sec.ppc}

The \emph{spatial sensitivity} and \emph{spatial smoothness} of a learnt representation may affect transfer performance on downstream tasks requiring dense prediction. The former measures the ability to discriminate spatially close pixels, needed for accurate prediction in boundary areas where labels change. The latter property encourages spatially close pixels to be similar, which can aid prediction in areas that belong to the same label. The \emph{PixContrast} method described in the last subsection only encourages the learnt representation to be \emph{spatially sensitive}. In the following, we present a new pixel-level pretext task which additionally introduces \emph{spatial smoothness} in the representation learning.

This new pretext task involves two critical components. The first is a pixel propagation module, which filters a pixel's features by propagating the features of similar pixels to it. This propagation has a feature denoising/smoothing effect on the learned representation that leads to more coherent solutions among pixels in pixel-level prediction tasks. The second component is an asymmetric architecture design where one branch produces a regular feature map and the other branch incorporates the pixel-propagation module. The pretext task seeks consistency between the features from the two branches without considering negative pairs. On the one hand, this design maintains the \emph{spatial sensitivity} of the learnt representation to some extent, thanks to the regular branch. On the other hand, while the performance of contrastive learning is known to be heavily affected by the treatment of negative pairs~\cite{he2019moco,chen2020simclr}, the asymmetric design enables the representation learning to rely only on consistency between positive pairs, without facing the issue of carefully tuning negative pairs~\cite{grill2020byol}. We refer to this pretext task as \emph{pixel-to-propagation consistency (PPC)} and describe these primary components in the following.

\begin{figure}
    \centering
    \includegraphics[width=0.8\linewidth]{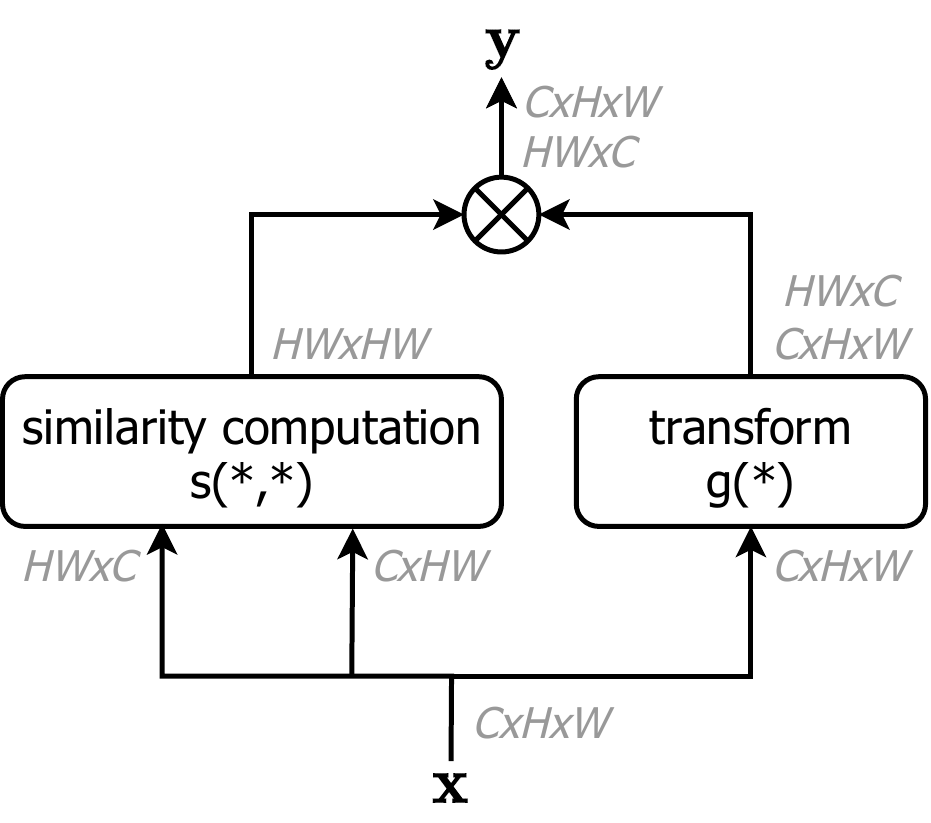}
    \caption{Illustration of the pixel propagation module (\emph{PPM}). The input and output resolutions of each computation block are included.}
    \label{fig:ppm}
\end{figure}

\paragraph{Pixel Propagation Module} For each pixel feature $\mathbf{x}_i$, the pixel propagation module computes its smoothed transform $\mathbf{y}_i$ by propagating features from all pixels $\mathbf{x}_j$ within the same image $\Omega$ as
\begin{equation}
\label{eq.ppm}
    \mathbf{y}_i = \Sigma _{j \in \Omega} s(\mathbf{x}_i, \mathbf{x}_j)\cdot g (\mathbf{x}_j),
\end{equation}
where $s(\cdot, \cdot)$ is a similarity function defined as
\begin{equation}
\label{eq.sim}
s(\mathbf{x}_i, \mathbf{x}_j) = \left (\max(\text{cos}(\mathbf{x}_i, \mathbf{x}_j), 0) \right)^\gamma,
\end{equation}
with $\gamma$ being an exponent to control the sharpness of the similarity function and is set by default to $2$; $g (\cdot)$ is a transformation function that can be instantiated by $l$ linear layers with a batch normalization and a ReLU layer between two successive layers. When $l=0$, $g (\cdot)$ is an identity function and Eq.~(\ref{eq.ppm}) will be a non-parametric module. Empirically, we find that all of $l=\{0,1,2\}$ perform well, and we set $l=1$ by default as its results are slightly better. Figure~\ref{fig:ppm} illustrates the proposed pixel propagation module.

\paragraph{Pixel-to-Propagation Consistency Loss} In the asymmetric architecture design, there are two different encoders: a regular encoder with the pixel propagation module applied afterwards to produce smoothed features, and a momentum encoder without the propagation module. The two augmentation views both pass through the two encoders, and the features from different encoders are encouraged to be consistent:
\begin{equation}
    \mathcal{L}_\text{PixPro} = -\text{cos}(\mathbf{y}_i, \mathbf{x}_j') - \text{cos}( \mathbf{y}_j, \mathbf{x}_i') ,
\end{equation}
where $i$ and $j$ are a positive pixel pair from two augmentation views according to the assignment rule in Eq.~(\ref{eq.assign_rule}); $\mathbf{x}_i'$ and $\mathbf{y}_i$ are pixel feature vectors of the momentum encoder and the propagation encoder, respectively. This loss is averaged over all positive pairs for each image, and then further averaged over images in a mini-batch to drive the representation learning.

\paragraph{Comparison to \emph{PixContrast}}
The overall architecture of the pixel-to-propagation consistency (\emph{PPC}) method is illustrated in Figure~\ref{fig:pix_architecture}. Compared to the \emph{PixContrast} method described in Section~\ref{sec.baseline} (see the blue-color loss in Figure~\ref{fig:pix_architecture}), there are two differences: the introduction of a pixel propagation module (\emph{PPM}), and replacement of the contrastive loss by a consistency loss. Table~\ref{table.pixpro_ablate}(c) and~\ref{table.pixctr_pixpro} show that both changes are critical for the feature transfer performance.

\paragraph{Computation complexity} The proposed \emph{PixContrast} and \emph{PixPro} approaches adopt the same data loader and backbone architectures as those of the instance discrimination based representation learning methods. There computation complexity in pre-training is thus similar as that of the counterpart instance-level method (i.e. BYOL~\cite{grill2020byol}): 8.6G vs. 8.2G FLOPs using a ResNet-50 backbone architecture, where the head and loss contribute about 0.4G FLOPs overhead. 

\subsection{Aligning Pre-training to Downstream Networks}

Previous visual feature pre-training methods are generally limited to classification backbones. For supervised pre-training, i.e. by the ImageNet image classification pretext task, the standard practice is to transfer only the pre-trained backbone features to downstream tasks. The recent unsupervised pre-training methods have continued this practice. One reason is that the pre-training methods operate at the instance level, making them less compatible with the dense prediction required in head networks for downstream tasks.

In contrast, the fine-grained spatial inference of pixel-level pretext tasks more naturally aligns with dense downstream tasks. To examine this, we consider an object detection method, FCOS~\cite{tian2019fcos}, for dense COCO detection. FCOS~\cite{tian2019fcos} applies a feature pyramid network (FPN) from P3 ($8\times$ down-sampling) to P7 ($128\times$ down-sampling)~\cite{FPN}, followed by two separate convolutional head networks (shared for all pyramidal levels) on the output feature maps of a ResNet backbone to produce classification and regression results.

We adopt the same architecture from the input image until the third $3\times 3$ convolutional layer in the head. In FPN, we involve feature maps from P3 to P6, with P7 omitted because the resolution is too low. A pixel propagation module (\emph{PPM}) with shared weights and the \emph{pixel-to-propagation consistency} (\emph{PPC}) loss described in Section~\ref{sec.ppc} are applied on each pyramid level. The final loss is first averaged at each pyramidal level and then averaged over all the pyramids.

Pre-training the FPN layers and the head networks used for downstream tasks can generally improve the transfer accuracy, as shown in Tables~\ref{table.head_pretrain} and \ref{table.semi_supervised}.

\subsection{Combined with Instance Contrast}

The presented pixel-level pretext tasks adopt the same data loader and encoders as in state-of-the-art instance-level discrimination methods~\cite{he2019moco,grill2020byol}, with two augmentation views sampled from each image and fed into backbone encoders. Hence, our pixel-level methods can be conveniently combined with instance-level pretext tasks, by sharing the same data loader and backbone encoders, with little pre-training overhead.

Specifically, the instance-level pretext task is applied on the output of the \emph{res5} stage, using projection heads that are independent of the pixel-level task. Here, we use a popular instance-level method, SimCLR~\cite{chen2020simclr}, with a momentum encoder to be aligned with the pixel-level pretext task. In this combination, the two losses from the pixel-level and instance-level pretext tasks are balanced by a multiplicative factor $\alpha$ (set to 1 by default), as
\begin{equation}
    \mathcal{L} = \mathcal{L}_\text{PixPro} + \alpha \mathcal{L}_\text{inst}.
\end{equation}

In general, the two tasks are complementary to each other: a pixel-level pretext task learns representations good for spatial inference, while an instance-level pretext task is good for learning categorization representations. Table~\ref{table.pix_inst} shows that an additional instance-level contrastive loss can significantly improve ImageNet-1K linear evaluation, indicating that a better categorization representation is learnt. Likely because of better categorization ability, it achieves noticeably improved transfer accuracy on the downstream task of FCOS~\cite{tian2019fcos} object detection on COCO (about 1 mAP improvement).

\section{Experiments}

\subsection{Pre-training Settings}

\begin{table*}[t]
\begin{center}
\begin{tabular}{l|c|ccc|ccc|ccc|c}
    \Xhline{2\arrayrulewidth}
        \multirow{2}{*}{Method} & \multirow{2}{*}{\#. Epoch} & \multicolumn{3}{c|}{Pascal VOC (R50-C4)} & \multicolumn{3}{c|}{COCO (R50-FPN)} & \multicolumn{3}{c|}{COCO (R50-C4)} & Cityscapes (R50)\\
         & & AP & $\text{AP}_\text{50}$ & $\text{AP}_\text{75}$ & mAP & $\text{AP}_\text{50}$ & $\text{AP}_\text{75}$ & mAP & $\text{AP}_\text{50}$ & $\text{AP}_\text{75}$ & mIoU\\
        \hline
        scratch & - & 33.8 & 60.2 & 33.1 & 32.8 & 51.0 & 35.3 & 26.4 & 44.0 & 27.8 & 65.3 \\
        supervised & 100 & 53.5 & 81.3 & 58.8 & 39.7 & 59.5 & 43.3 & 38.2 & 58.2 & 41.2 & 74.6 \\
        \hline
        MoCo~\cite{he2019moco} & 200 & 55.9 & 81.5 & 62.6 & 39.4 & 59.1 & 43.0 & 38.5 & 58.3 & 41.6 & 75.3 \\
        SimCLR~\cite{chen2020simclr} & 1000 & 56.3 & 81.9 & 62.5 & 39.8 & 59.5 & 43.6 & 38.4 & 58.3 & 41.6 & 75.8 \\
        MoCo v2~\cite{chen2020mocov2} & 800 & 57.6 & 82.7 & 64.4 & 40.4 & 60.1 & 44.3 & 39.5 & 59.0 & 42.6 & 76.2 \\
        InfoMin~\cite{tian2020infomin} & 200 & 57.6 & 82.7 & 64.6 & 40.6 & 60.6 & 44.6 & 39.0 & 58.5 & 42.0 & 75.6 \\
        InfoMin~\cite{tian2020infomin} & 800 & 57.5 & 82.5 & 64.0 & 40.4 & 60.4 & 44.3 & 38.8 & 58.2 & 41.7 & 75.6 \\
        \hline
        \emph{PixPro} (ours)  & 100 & 58.8 & 83.0 & 66.5 & 41.3 & 61.3 & 45.4 & 40.0 & 59.3 & 43.4 & 76.8 \\
        \emph{PixPro} (ours)  & 400 & \textbf{60.2} & \textbf{83.8} & \textbf{67.7} & \textbf{41.4} & \textbf{61.6} & \textbf{45.4} & \textbf{40.5} & \textbf{59.8} & \textbf{44.0} & \textbf{77.2} \\
        \Xhline{2\arrayrulewidth}
    \end{tabular}
\end{center}
\caption{Comparing the proposed pixel-level pre-training method, \emph{PixPro}, to previous supervised/unsupervised pre-training methods. For Pascal VOC object detection, a Faster R-CNN (R50-C4) detector is adopted for all methods. For COCO object detection, a Mask R-CNN detector (R50-FPN and R50-C4) with $1\times$ setting is adopted for all methods. For Cityscapes semantic segmentation, an FCN method (R50) is used. Only a pixel-level pretext task is involved in \emph{PixPro} pre-training. For Pascal VOC (R50-C4), COCO (R50-C4) and Cityscapes (R50), a regular backbone network of R50 with output feature map of C5 is adopted for \emph{PixPro} pre-training. For COCO (R50-FPN), an FPN network with $P_3$-$P_6$ feature maps is used. Note that InfoMin~\cite{tian2020infomin} reports results for only its 200 epoch model, so we reproduce it with longer training lengths, where saturation is observed.}
\label{tab:system}
\end{table*}

\paragraph{Datasets}
We adopt the widely used ImageNet-1K~\cite{deng2009imagenet} dataset for feature pre-training, which consists of $\sim$1.28 million training images.

\paragraph{Architectures} Following recent unsupervised methods~\cite{he2019moco,grill2020byol}, we adopt the ResNet-50~\cite{he2016resnet} model as our backbone network. The two branches use different encoders, with one using a regular backbone network and a regular projection head, and the other using the momentum network with a moving average of the parameters of the regular backbone network and the projection head. The proposed pixel propagation module (\emph{PPM}) is applied on the regular branch.The FPN architecture with P3-P6 feature maps are also tested in some experiments.

\paragraph{Data Augmentation} In pre-training, the data augmentation strategy follows~\cite{grill2020byol}, where two random crops from the image are independently sampled and resized to $224\times 224$ with a random horizontal flip, followed by color distortion, Gaussian blur, and a solarization operation. We skip the loss computation for cropped pairs with no overlaps, which compose only a small fraction of all the cropped pairs.

\paragraph{Optimization} We vary the training length from 50 to 400 epochs, and use 100-epoch training in our ablation study. The \emph{LARS} optimizer with a cosine learning rate scheduler and a base learning rate of 1.0 is adopted in training, where the learning rate is linearly scaled with the batch size as $\text{lr}=\text{lr}_\text{base} \times \text{\#}\text{bs}/256$. Weight decay is set to 1e-5. The total batch size is set to 1024, using 8 V100 GPUs. For the momentum encoder, the momentum value starts from 0.99 and is increased to 1, following~\cite{grill2020byol}. Synchronized batch normalization is enabled during training.

\subsection{Downstream Tasks and Settings}

We evaluate feature transfer performance on four downstream tasks: object detection on Pascal VOC~\cite{everingham2010pascalvoc}, object detection on COCO~\cite{lin2014microsoft}, semantic segmentation on Cityscapes~\cite{cordts2016cityscapes}, and semi-supervised object detection on COCO~\cite{sohn2020simple}. In some experiments, we also report the ImageNet-1K~\cite{deng2009imagenet} linear evaluation performance for reference.

\paragraph{Pascal VOC Object Detection} We strictly follow 
the setting introduced in~\cite{he2019moco}, namely a Faster R-CNN detector~\cite{ren2015faster} with the ResNet50-C4 backbone, which uses the \emph{conv4} feature map to produce object proposals and uses the \emph{conv5} stage for proposal classification and regression. In fine-tuning, we synchronize all batch normalization layers and optimize all layers. In testing, we report AP, AP50 and AP75 on the {\fontfamily{pcr}\selectfont test2007} set. Detectron2~\cite{wu2019detectron2} is used as the code base.

\paragraph{COCO Object Detection and Instance Segmentation} We adopt the Mask R-CNN detector with ResNet50-FPN and ResNet50-C4~\cite{Mask-rcnn,FPN} backbones, respectively. In optimization, we follow the $1\times$ settings, with all batch normalization layers synchronized and all layers fine-tuned~\cite{he2019moco}. We adopt Detectron2~\cite{wu2019detectron2} as the code base for these experiments.

We also consider other detectors with fully convolutional architectures, e.g., FCOS~\cite{tian2019fcos}. For these experiments, we follow the $1\times$ settings and utilize the mmdetection code base~\cite{chen2019mmdetection}.

\paragraph{Cityscapes Semantic Segmentation} We follow the settings of MoCo~\cite{he2019moco}, where an FCN-based structure is used~\cite{long2015fully}. The FCN network consists of a ResNet-50 backbone with $3\times 3$ convolution layers in the \emph{conv5} stage of dilation 2 and stride 1,
followed by two $3\times 3$ convolution layers of 256 channels and dilation 6. The classification is obtained by an additional $1\times 1$ convolutional layer.

\paragraph{Semi-Supervised Object Detection} We also examined semi-supervised learning for object detection on COCO. For this, a small fraction (1\%-10\%) of images randomly sampled from the training set is assigned labels and used in fine-tuning. The results of five random trials are averaged for each method.

\paragraph{ImageNet-1K Linear Evaluation} In this task, we fix the pretrained features and only fine-tune one additional linear classification layer, exactly following the settings of MoCo~\cite{he2019moco}. We report these results for reference.

\setlength{\tabcolsep}{3pt}
\renewcommand{\arraystretch}{1.1}
\begin{table}[t]
\small
\begin{center}
\begin{tabular}{l|ccc|c}
\hline
\multirow{2}{*}{parameters} &
\multicolumn{3}{c|}{Pascal VOC} & COCO \\
\cline{2-5}
& AP & AP$_\text{50}$ & AP$_\text{70}$ & mAP \\
\hline
\multicolumn{5}{l}{\textbf{(a)} distance threshold $\mathcal{T}$ using $\text{C}_5$ ($7\times7$)}\\
\hline
$\mathcal{T}=0.35$ & 58.3 & 82.1 & 65.8 & 39.5 \\
$\mathcal{T}=0.7(*)$ & \textbf{58.8} & \textbf{83.0} & \textbf{66.5} & \textbf{40.8} \\
$\mathcal{T}=1.4$ & 56.8 & 82.0 & 63.3 & 39.5 \\
$\mathcal{T}=2.8$ & 56.5 & 81.7 & 63.4 & 39.1 \\
\hline
\multicolumn{5}{l}{\textbf{(b)} distance threshold $\mathcal{T}$ using $\text{P}_3$ ($28\times 28$)}\\
\hline
$\mathcal{T}=0.35$   & \textbf{58.1} & \textbf{83.0} & \textbf{64.7} & \textbf{40.8} \\
$\mathcal{T}=0.7$ & 57.6 & \textbf{83.0} & 63.6 & \textbf{40.8} \\
$\mathcal{T}=1.4$    & 56.8 & 82.7 & 63.1 & 40.6 \\
$\mathcal{T}=2.8$    & 56.1 & 82.4 & \textbf{64.7} & 40.2 \\
\hline
\multicolumn{5}{l}{ \textbf{(c)} sharpness exponent $\gamma$} \\
\hline
$\gamma=0.5$ & 57.9 & 82.5 & 64.5 & 39.7 \\
$\gamma=1$ & 58.7 & 83.0 & 65.5 & 40.1 \\
$\gamma=2(*)$ & \textbf{58.8} & \textbf{83.0} & \textbf{66.5} & \textbf{40.8} \\
$\gamma=4$ & 58.0 & 82.4 & 64.7 & 40.0 \\
$\gamma=8$ & 57.8 &	82.5 & 64.4 & 39.9 \\
\hline
\multicolumn{5}{l}{ \textbf{(d)} number of transformation layers in $g(\cdot)$} \\
\hline
$l=0$ & 58.6 & 82.9	& 65.4 & 39.4 \\
$l=1(*)$ & 58.8 & 83.0 & \textbf{66.5} & \textbf{40.8} \\
$l=2$ & \textbf{58.9} & \textbf{83.1} & 66.3 & 40.3 \\
$l=3$ & 58.3 & 82.5 & 65.0 & 40.1 \\
\hline
\multicolumn{5}{l}{ \textbf{(e)} output resolution} \\
\hline
$\text{C}_5$ ($7\times 7*$)  & \textbf{58.8} & \textbf{83.0} & \textbf{66.5} & 40.8  \\
$\text{P}_4$ ($14\times 14$) & 56.7 & 82.7 & 63.6 & 40.9 \\
$\text{P}_3$ ($28\times 28$) & 57.6 & \textbf{83.0} & 63.6 & 40.8 \\
$\text{P}_3$-$\text{P}_6$    & 55.8 & 82.5 & 62.1 & \textbf{41.3} \\
\hline
\multicolumn{5}{l}{\textbf{(f)} training length} \\
\hline
50 epoch  & 57.2 & 82.4 & 63.4 & 39.7 \\
100 epoch(*) & 58.8 & 83.0 & 66.5 & 40.8 \\
200 epoch & 59.5 & 83.5	& 66.9 & 40.8 \\
400 epoch & \textbf{60.2} & \textbf{83.8}	& \textbf{67.7} & \textbf{41.0} \\
\hline
\end{tabular}
\end{center}
\caption{Ablation studies on hyper-parameters for the proposed \emph{PixPro} method. Rows with (*) indicate default values.}
\label{table.pixpro_ablate}
\end{table}

\subsection{Main Transfer Results}

Table~\ref{tab:system} compares the proposed method to previous state-of-the-art unsupervised pre-training approaches on 4 downstream tasks, which all require dense prediction. Our \emph{PixPro} achieves 60.2 AP, 41.4 / 40.5 mAP and 77.2 mIoU on Pascal VOC object detection (R50-C4), COCO object detection (R50-FPN / R50-C4) and Cityscapes semantic segmentation (R50). It outperforms the previous best unsupervised methods by 2.6 AP on Pascal VOC, 0.8 / 1.0 mAP on COCO and 1.0 mIoU on Cityscapes.

\subsection{Ablation Study}

We conduct the ablation study using the Pascal VOC (R50-C4) and COCO object detection (R50-FPN) tasks. In some experiments, the results of the FCOS detector on COCO and semi-supervised results are included.

\paragraph{Hyper-Parameters for \emph{PixPro}} Table~\ref{table.pixpro_ablate} examines the sensitivity to hyper-parameters of \emph{PixPro}. For the ablation of each hyper-parameter, we fix all other hyper-parameters to the following default values: feature map of $\text{C}_5$, distance threshold $\mathcal{T}=0.7$, sharpness exponent $\gamma=2$, number of transformation layers in the pixel-to-propagation module $l=1$, and training length of 100 epochs.

Table~\ref{table.pixpro_ablate} (a-b) ablates distance thresholds using the feature maps of $\text{C}_5$ and $\text{P}_3$. For both, $\mathcal{T}=0.7$ yields good performance. The results are more stable for $\text{P}_3$, thanks to its larger resolution.

Table~\ref{table.pixpro_ablate} (c) ablates the sharpness exponent $\gamma$, where $\gamma=2$ shows the best results. A similarity function that is too smooth or too sharp harms transfer performance.

Table~\ref{table.pixpro_ablate} (d) ablates the number of transformation layers in $g(\cdot)$, where $l=1$ shows slightly better performance than others. Note that $l=0$, which has no learnable parameters in the \emph{pixel-propagation module (PPM)}, also performs reasonably well, while removal of the \emph{PPM} module results in model collapse. The smoothness operation in the \emph{PPM} module introduces asymmetry with respect to the other regular branch, and consequently avoids collapse~\cite{grill2020byol}.

Table~\ref{table.pixpro_ablate} (e) ablates the choice of feature maps. It can be seen that using the higher resolution feature maps of $\text{P}_3$ and $\text{P}_4$ performs similarly well to using $\text{C}_5$. Using all $\text{P}_3$-$\text{P}_6$ feature maps noticeably improves the transfer accuracy on COCO object detection, but is inferior to others on Pascal VOC object detection. As the Pascal VOC dataset uses a ResNet-C4 backbone and the COCO dataset uses a ResNet-FPN backbone, this result suggests that consistent architecture between pre-training and the downstream task may give better results.

Table~\ref{table.pixpro_ablate} (f) ablates the effects of training length. Increasing the training length generally results in better transfer performance. Our maximal training length is 400. Compared to 200 epoch training, it brings a 0.7 AP gain on Pascal VOC, while almost saturating on COCO. Results on longer training will be examined in our future work.

\paragraph{Comparison of \emph{PixPro} and \emph{PixContrast}} Table~\ref{table.pixctr_pixpro} ablates the transfer performance of \emph{PixContrast} with varying $\tau$ and with/without the pixel propagation module (\emph{PPM}). It also includes the results of the \emph{PixPro} method with/without the \emph{PPM}. It can be seen that while the \emph{PixContrast} method achieves reasonable transfer performance, the \emph{PixPro} method is better, specifically 0.7 AP and 2.0 mAP better than the \emph{PixContrast} approach on Pascal VOC and COCO, respectively.

Including the pixel-propagation module (\emph{PPM}) leads to inferior performance for the \emph{PixContrast} method, likely because of over-smoothing. In contrast, for \emph{PixPro}, adding \emph{PPM} improves transfer performance by 0.8 AP on Pascal VOC and 1.1 mAP on COCO, as well as avoids the use of hyper-parameter $\tau$. Note while directly removing \emph{PPM} will result in model collapse, we add a linear transformation layer to avoid such collapse issue. Also note that the benefit of this spatial smoothness in representation learning is also evidenced in Table~\ref{table.pixpro_ablate}(c), where a similarity function that is too smooth or too sharp harms transfer performance.

\begin{table}[t]
\begin{center}
    \begin{tabular}{c|c|c|ccc|c}
\hline
\multirow{2}{*}{method} &
\multirow{2}{*}{\emph{PPM}} & \multirow{2}{*}{$\tau$} &
\multicolumn{3}{c|}{Pascal VOC} & COCO \\
\cline{4-7}
&  &  & AP & $\text{AP}_\text{50}$ & $\text{AP}_\text{75}$ & mAP \\
\hline
\multirow{6}{*}{\emph{PixContrast}} & & 0.1 & 54.7 & 79.9 & 61.2 & 38.0 \\
 & &  0.2 & 57.1 & 81.7 & 63.3  & 38.6 \\
 & & 0.3 & \textbf{58.1} & \textbf{82.4} & \textbf{64.5} & \textbf{38.8} \\
\cline{2-7}
& \checkmark & 0.1 & 52.7 & 78.8 & 57.6 & 37.4 \\
 & \checkmark &  0.2 & 53.0 & 79.1 & 58.1 & 37.3 \\
 & \checkmark & 0.3 & 52.9 & 78.8 & 58.3 & 37.5 \\
\hline
 \multirow{2}{*}{\emph{PixPro}} &  & - & 58.0 & 82.6 & 65.6 & 39.7 \\
 & \checkmark & - & \textbf{58.8} & \textbf{83.0} & \textbf{66.5} & \textbf{40.8} \\
 \hline
\end{tabular}
\end{center}
\caption{Comparison of the \emph{PixContrast} and \emph{PixPro} methods. 100 epoch pre-training is adopted for all experiments.}
\label{table.pixctr_pixpro}
\end{table}

\begin{comment}
\begin{table}[t]
\begin{center}
    \begin{tabular}{c|c|cc|c|c}
\hline
\multirow{2}{*}{method} &
\multirow{2}{*}{\emph{PPM}} & \multicolumn{2}{c|}{negative scope} &
VOC & COCO \\
\cline{3-6}
&  & within im & cross im & AP & mAP \\
\hline
\multirow{3}{*}{\emph{PixContrast}} &\multirow{3}{*}{none} & \checkmark & & & \\
 & &  & \checkmark & & \\
 & & \checkmark & \checkmark & & \\
\hline
\multirow{6}{*}{\emph{ProContrast}} &\multirow{3}{*}{2-branch} & \checkmark & & & \\
 & &  & \checkmark & & \\
 & & \checkmark & \checkmark & & \\
 \cline{2-6}
  &\multirow{3}{*}{1-branch} & \checkmark & & & \\
 & &  & \checkmark & & \\
 & & \checkmark & \checkmark & & \\
\hline
 \emph{PixPro} & 1-branch &  & & & \\
 \hline
\end{tabular}
\end{center}
\caption{Comparison of the \emph{PixContrast} and \emph{PixPro} methods.}
\label{table.pixctr_pixpro}
\end{table}
\end{comment}

\paragraph{Combined with Instance-Level Contrastive Methods} Table~\ref{table.pix_inst} ablates the effects of combining the proposed \emph{PixPro} method with an instance-level pretext task (SimCLR*) for representation learning. The combination requires marginal added computation due to sharing of the data loader and encoders. It can be seen that an additional instance-level pretext task can significantly improve the linear evaluation accuracy on ImageNet-1K, while the transfer accuracy on COCO (mask R-CNN R50-FPN) and Pascal VOC is maintained. We also observe noticeable transfer improvements of 1.2 mAP on some tasks, e.g., FCOS~\cite{tian2019fcos} on COCO, as shown in Table~\ref{table.head_pretrain}.

\begin{table}[t]
\begin{center}
    \begin{tabular}{c|c|c|c|c}
\hline
\multirow{2}{*}{\makecell{\emph{PixPro}\\(pixel)}} &
\multirow{2}{*}{\makecell{SimCLR*\\(instance)}} &
VOC & COCO & ImageNet\\
\cline{3-5}
 &  & AP & mAP & top-1 acc \\
\hline
\checkmark &   & 58.8 & 40.8 & 55.1 \\
 & \checkmark & 53.4 & 40.5 & 65.4 \\
\checkmark & \checkmark & 58.7 & 40.9 & 66.3 \\
 \hline
\end{tabular}
\end{center}
\caption{Transfer performance of combining a pixel-level and an instance-level method. ``SimCLR*'' denotes a variant of SimCLR with the same encoders as our pixel-level approach. 100 epoch pre-training is adopted for all experiments.}
\label{table.pix_inst}
\end{table}

\paragraph{Effects of Head Network Pre-Training} Table~\ref{table.head_pretrain} ablates head network pre-training (or using an architecture more similar to that in the fine-tuning task) on COCO object detection. For COCO object detection, we use the FCOS detector, which is fully convolutional. We evaluate the transfer performance with an additional FPN architecture, a head network of three successive convolutional layers. It can be seen that more pre-training layers lead to better transfer accuracy on downstream tasks.

\begin{table}[t]
\begin{center}
    \begin{tabular}{c|c|c|ccc}
\hline
\multirow{2}{*}{+FPN} & \multirow{2}{*}{+head} &
\multirow{2}{*}{+instance} & \multicolumn{3}{c}{COCO (FCOS)} \\
\cline{4-6}
&  & & mAP & AP$_\text{50}$ & AP$_\text{75}$ \\
\hline
 &  &  & 37.8 & 56.2 & 40.6 \\
\checkmark &  & & 38.1 & 56.7 & 41.2 \\
\checkmark & \checkmark & & 38.6 & 57.3 & 41.5\\
\checkmark & \checkmark & \checkmark & \textbf{39.8} & \textbf{58.4} & \textbf{42.7} \\
\hline
\end{tabular}
\end{center}
\caption{FPN and head pre-training with transfer to COCO using an FCOS detector~\cite{tian2019fcos}. 100 epoch pre-training is adopted for all experiments.}
\label{table.head_pretrain}
\end{table}

\begin{table}[t]
\begin{center}
    \begin{tabular}{l|c|cc}
\hline
\multirow{2}{*}{arch.} & \multirow{2}{*}{\makecell{+COCO\\pre-train}} &  \multicolumn{2}{c}{\makecell{mask\\R-CNN}} \\
\cline{3-4}
&  & 1\% & 10\% \\
\hline
supervised & & 10.4 & 20.4   \\
\hline
MoCo~\cite{he2019moco} & & 10.9 & 23.8  \\
MoCo v2~\cite{chen2020mocov2} & & 10.9 & 23.9 \\
InfoMin~\cite{tian2020infomin} & & 10.6 & 24.5  \\
\hline
C5 backbone & & 13.2 & 25.9 \\
FPN & & 14.1 & 26.6 \\
FPN & \checkmark  & \textbf{14.8} & \textbf{26.8}\\
\hline
\end{tabular}
\end{center}
\caption{Semi-supervised object detection on COCO. 100-epoch pre-training is adopted for our method, and other methods use the models with their longest training.}
\label{table.semi_supervised}
\end{table}

\paragraph{Semi-Supervised Object Detection Results} Table~\ref{table.semi_supervised} shows the semi-supervised results using 1\% and 10\% of labeled data on COCO. The Mask R-CNN (R50-FPN) detector is tested. Our best pre-training models perform significantly better than previous instance-level supervised/unsupervised methods. The gains are +3.9 mAP and +2.3 mAP using 1\% and 10\% training data, respectively.

The results indicate the advantage of aligning networks between the pre-training and downstream tasks. Including the additional FPN layers in pre-training brings +0.9 and +0.7 mAP gains over the method which pre-trains only the plain backbone network (14.1 and 26.6 vs. 13.2 and 25.9). 

We also include an additional pre-training stage on COCO using the proposed pixel-level pretext task by 120 epochs, after the ImageNet-1K pre-training. It leads to additional +0.7 mAP gains and +0.2 mAP gains when 1\% and 10\% training data are used, respectively. The additional pre-training directly on down-stream unlabelled data may benefit the learning when only scarce labeled data is available.

\section{Conclusion}

This paper explores the use of pixel-level pretext tasks for learning dense feature representations. We first directly apply contrastive learning at the pixel level, leading to reasonable transfer performance on downstream tasks requiring dense prediction. We additionally propose a \emph{pixel-to-propagation consistency} task which introduces certain smoothness priors in the representation learning process and does not require processing of negative samples. This method, named \emph{PixPro}, achieves 60.2 AP and 41.4 / 40.5 mAP accuracy when the learnt representation is transferred to the downstream tasks of Pascal VOC (Faster R-CNN R50-C4) and COCO object detection (mask R-CNN R50-FPN / R50-C4), which are 2.6 AP and 0.8 / 1.0 mAP better than the previous best supervised/unsupervised pre-training methods. These results demonstrate the strong potential of defining pretext tasks at the pixel level, and suggest a new path forward in unsupervised visual representation learning. As a generic pretext task for learning stronger representations on single images, the proposed approach is also applicable to visual representation learning on videos and multi-modality signals.

{\small
\bibliographystyle{ieee_fullname}
\bibliography{egbib}
}

\end{document}